\documentclass[11pt]{article}
\usepackage{amsmath,amssymb,amsthm}
\usepackage{geometry}
\usepackage{bm}
\usepackage{graphicx}
\usepackage[hidelinks,hypertexnames=false]{hyperref}
\usepackage{float}
\usepackage{microtype}
\newtheorem{proposition}{Proposition}

\title{Predictive Assistance and the Temporal Dynamics of Exploratory Compression}

\author{
Balaraju Battu$^{1,2}$ \\
{\small $^{1}$European University Institute, Florence, Italy} \\
{\small $^{2}$New York University Abu Dhabi, UAE}
}
\date{\today}

\begin{document}
\maketitle

\begin{abstract}
When does predictive assistance change not only current problem solving, but also later exploratory capacity? This paper studies a specific temporal mechanism: assistance that stabilizes a candidate trajectory before self-generated exploration has diversified. Classical accounts of skill acquisition place exploratory variation before compression, so that efficient representations emerge from prior search. Predictive systems can reverse this order by supplying a solution path at the beginning of deliberation. We develop a minimal dynamical model with two endogenous states: accumulated self-generated exploratory breadth and local compression around an assisted trajectory. The model compares immediate and delayed assistance schedules with identical duration and intensity. Immediate assistance produces greater post-assistance compression when exploratory breadth forms faster than compressed structure relaxes. It then yields lower exploratory variance, lower strategy entropy, and slower recovery after support is withdrawn, even though the two schedules provide the same total assistance and leave the same accumulated breadth at the comparison time. The result is conditional rather than anti-assistance: when structural relaxation is sufficiently rapid, the timing disadvantage disappears, and in safety-critical tasks compression may be desirable. The framework yields a direct experimental test comparing equal-dose immediate and delayed assistance followed by unaided transfer. It therefore recasts predictive assistance as a problem of temporal ordering: stabilization can have different long-run effects depending on whether it arrives before or after exploratory diversification.
\end{abstract}

\section{Introduction}

Classical cognitive science describes problem solving as search through structured problem spaces rather than exhaustive optimization \cite{newell1994unified,simon1955behavioral,gigerenzer1999simple}. Early attempts, failed decompositions, competing hypotheses, and representational revisions expose a learner to multiple routes through a problem. With experience, this initially broad search becomes compressed into efficient chunks, stable representations, and reusable procedures \cite{chase1973perception,newell1972human}. In this classical sequence, exploration precedes compression.

Predictive artificial intelligence can reverse that order. A completion system can supply an outline before a writer has generated competing structures, a coding assistant can propose an implementation before alternative decompositions have been attempted, and a decision aid can stabilize a candidate interpretation before uncertainty has produced independent hypotheses. Such assistance may improve immediate performance. The question considered here is different: \emph{what happens to later exploratory capacity when predictive assistance stabilizes a problem-solving trajectory before self-generated exploration has diversified?}

The paper's answer is that early predictive stabilization can produce \emph{exogenous exploratory compression}. When a candidate trajectory is supplied while the learner's repertoire remains narrow, stabilization is concentrated around that trajectory. Repeated concentration increases local curvature, narrowing the range over which subsequent perturbations redirect attention. Because stabilized structure dissipates gradually, the resulting reduction in exploratory mobility can remain after assistance is removed. Assistance delivered after an initial exploratory interval acts on a broader repertoire and can therefore leave less residual compression, even when the duration and intensity of assistance are held fixed.

This claim is narrower than the proposition that AI generally damages cognition. Assistance can expose users to alternatives, provoke disagreement, or enable search that would otherwise be infeasible. Compression is also desirable when reliable convergence is the objective, as in time-critical or safety-critical procedures. The mechanism studied here concerns predictive systems that supply and repeatedly stabilize a candidate trajectory in tasks where broad early exploration contributes to later transfer, revision, or adaptation. Its central variable is temporal order: whether stabilization arrives before or after self-generated diversification.

Several literatures motivate this mechanism without yet providing a unified timing account. Work on desirable difficulty and productive failure shows that delay, partial failure, and self-generation can improve transfer even when direct instruction improves immediate performance \cite{bjork2011desirable,kapur2008productive}. Research on curiosity, spontaneous thought, and information seeking emphasizes the role of internally generated variation in revising representations \cite{gottlieb2013information,gruber2014states,christoff2016mind}. Studies of cognitive offloading and navigation similarly suggest that externally supplied trajectories can alter later independent search and cognitive mapping \cite{dahmani2020gps,macnamara2024ai,leon2024cognitive}. Human-AI interaction research has documented automation reliance and the difficulty of sustaining appropriate engagement with decision aids \cite{parasuraman2000model,buccinca2021trust,amershi2019guidelines}. The present contribution is to isolate a common dynamical mechanism: the same assistance can have different post-assistance effects depending on when it enters the exploratory process.

We formalize this claim with a deliberately minimal model. Exploratory breadth $B(t)$ records the share of a potential strategy repertoire generated through independent search. Compression $C(t)$ records additional curvature around a focal assisted trajectory. Assistance slows the formation of self-generated breadth and produces more compression when breadth is low. Local deliberation is represented as stochastic motion in a basin whose curvature is the sum of baseline stabilization and accumulated compression. This construction links compression to observable quantities: exploratory variance, differential entropy, and the time required to leave a stabilized region.

The model yields one main result. Compare an immediate assistance block with an equally long delayed assistance block over a common learning horizon. Both schedules contain the same amount of assisted and unassisted time, and in the baseline model both generate the same final amount of self-generated breadth. Nevertheless, immediate assistance leaves greater residual compression precisely when repertoire formation is faster than structural relaxation. This condition, $\rho>\delta$, identifies the domain in which stabilization-before-exploration has persistent consequences. After assistance is removed, the difference decays rather than remaining irreversible, producing delayed recovery and fading structural memory rather than equilibrium hysteresis.

The result generates a direct empirical design. Participants receive identical predictive guidance either at the start of a learning episode or after a fixed interval of independent exploration. Assistance is then withdrawn and exploratory behavior is measured on structurally related transfer tasks. The theory predicts lower strategy entropy, slower abandonment of the assisted trajectory after contradiction, and slower recovery across unaided trials in the immediate-assistance condition when the timing condition holds. A null post-withdrawal difference, or rapid convergence of the two groups, would reject the proposed mechanism in that domain.

The remainder of the paper develops the temporal distinction, presents the model and timing result, derives its behavioral implications, and specifies an empirical test and its boundary conditions.

\section{Exploration Before Compression}

Exploration and compression are not opposites. Efficient cognition requires both. Exploration distributes attention across possible representations, while compression stabilizes regularities discovered through that search. The theoretical issue is therefore not whether compression occurs, but how its temporal relation to exploration shapes the structure that remains available for later use.

Endogenous compression follows self-generated search. A learner first constructs, compares, or rejects multiple candidate strategies; repeated success then concentrates attention around a smaller set of efficient representations. Predictive assistance can instead generate exogenous compression by supplying a focal trajectory before this repertoire has formed. The resulting performance may be locally successful, yet success alone does not reveal whether the learner can later produce alternatives, revise the supplied representation, or transfer the underlying structure to a novel setting.

Consider an open-ended causal reasoning task. Without assistance, a learner may generate several causal graphs, design discriminating tests, revise an initial model after contradictory evidence, and only then converge. With immediate predictive assistance, a plausible graph and next experiment may be supplied at the outset. The learner can evaluate and use that proposal efficiently, but the proposal also becomes the initial organizing structure around which later evidence is interpreted. If the same guidance is delayed until after several self-generated models have been considered, it enters a different representational environment: the learner already possesses alternatives against which the supplied trajectory can be compared.

This distinction suggests three requirements for a timing theory. First, the model must represent exploratory diversification explicitly rather than infer it from elapsed time. Second, stabilization must be stronger when assistance arrives before diversification. Third, the post-assistance outcome must be measured after current support is removed; otherwise improved assisted performance and diminished independent exploration cannot be distinguished. The model below introduces the smallest state system satisfying these requirements.

\section{A Timing Model of Exploratory Compression}

\subsection{Exploratory breadth and assistance}

Let $B(t)\in[0,1]$ denote accumulated self-generated exploratory breadth. The value $B=0$ represents a repertoire in which no relevant alternatives have yet been independently generated, while $B=1$ represents saturation of the task-relevant repertoire. Let $A(t)\in[0,1]$ denote the degree to which the current problem-solving trajectory is externally stabilized. At $A=0$, search is independent. At $A=1$, the relevant trajectory is fully supplied or guided, so the current interval does not add self-generated breadth. Intermediate values represent partial stabilization.

Exploratory breadth evolves according to
\begin{equation}
\dot B(t)=\rho\,[1-A(t)]\,[1-B(t)],
\label{eq:breadth}
\end{equation}
where $\rho>0$ is the rate at which independent search expands the repertoire. The saturation term $1-B$ captures diminishing additions to breadth as more of the task-relevant strategy space has already been sampled. Equation~\eqref{eq:breadth} does not claim that assisted users are mentally inactive. It distinguishes self-generated diversification from evaluation or execution of an externally stabilized trajectory.

The assistance index can summarize more than one interface operation. Anticipatory completion supplies a candidate representation before search begins, while difficulty regulation can remove the unresolved steps that would otherwise elicit independent branching. These interventions differ operationally, but they belong in the present model only to the extent that they stabilize the trajectory and reduce self-generated diversification. Systems that actively generate competing hypotheses or require independent reconstruction can have a lower effective $A(t)$ even when they provide substantial computational support.

\subsection{Compression as local curvature}

Let $C(t)\geq0$ denote accumulated compression around a focal trajectory. Compression evolves according to
\begin{equation}
\dot C(t)=\alpha A(t)[1-B(t)]-\delta C(t),
\label{eq:compression}
\end{equation}
where $\alpha>0$ is the stabilizing strength of assistance and $\delta>0$ is the relaxation rate. The interaction $A(1-B)$ is the timing mechanism. Identical assistance has a stronger compressive effect when the self-generated repertoire is narrow. Once breadth is high, an externally supplied trajectory competes with established alternatives and contributes less additional concentration.

To connect this stock to exploratory behavior, consider a local approximation around the assisted trajectory $m$. Let $X_t$ denote the current attentional or representational state. In a focal basin,
\begin{equation}
U(x,t)=\frac{1}{2}[\lambda_0+C(t)](x-m)^2,
\label{eq:potential}
\end{equation}
where $\lambda_0>0$ is baseline curvature. Deliberation follows the stochastic dynamics
\begin{equation}
dX_t=-[\lambda_0+C(t)](X_t-m)\,dt+\sqrt{2D}\,dW_t,
\label{eq:attention}
\end{equation}
with $D>0$ denoting the intensity of endogenous exploratory perturbation. Compression is therefore literal local curvature rather than an independent label attached to the landscape.

For fixed $C$, Equation~\eqref{eq:attention} is an Ornstein-Uhlenbeck process with stationary variance
\begin{equation}
V(C)=\frac{D}{\lambda_0+C}.
\label{eq:variance}
\end{equation}
Define normalized exploratory responsiveness as
\begin{equation}
R(C)=\frac{V(C)}{V(0)}=\frac{\lambda_0}{\lambda_0+C}.
\label{eq:responsiveness}
\end{equation}
Both variance and responsiveness decline monotonically with compression. Endogenous perturbations remain present, but they move attention across a narrower region. The corresponding differential entropy is
\begin{equation}
h(C)=\frac{1}{2}\log\!\left(\frac{2\pi eD}{\lambda_0+C}\right),
\label{eq:entropy}
\end{equation}
which provides a direct bridge between the latent state $C$ and observable diversity in exploratory trajectories.

\section{Temporal Ordering and Delayed Recovery}

\subsection{Equal-dose immediate and delayed assistance}

The central comparison holds total assistance fixed. Let the learning horizon be $[0,2L]$. The immediate schedule supplies full assistance during the first block,
\begin{equation}
A_E(t)=
\begin{cases}
1, & 0\leq t<L,\\
0, & L\leq t\leq2L,
\end{cases}
\end{equation}
whereas the delayed schedule reverses the order,
\begin{equation}
A_D(t)=
\begin{cases}
0, & 0\leq t<L,\\
1, & L\leq t\leq2L.
\end{cases}
\end{equation}
The schedules contain identical assistance intensity and duration. They differ only in whether stabilization occurs before or after an interval of self-generated exploration.

\begin{proposition}[Temporal ordering]
Suppose $B(0)=B_0\in[0,1)$ and $C(0)=0$. At time $2L$, the immediate and delayed schedules generate the same exploratory breadth,
\begin{equation}
B_E(2L)=B_D(2L)=1-(1-B_0)e^{-\rho L},
\end{equation}
but their residual compression levels are
\begin{align}
C_E(2L)&=\frac{\alpha(1-B_0)}{\delta}(1-e^{-\delta L})e^{-\delta L},\\
C_D(2L)&=\frac{\alpha(1-B_0)}{\delta}(1-e^{-\delta L})e^{-\rho L}.
\end{align}
Consequently,
\begin{equation}
C_E(2L)>C_D(2L)\quad\Longleftrightarrow\quad \rho>\delta.
\label{eq:timingcondition}
\end{equation}
When exploratory breadth forms faster than compressed structure relaxes, assistance supplied before diversification leaves greater post-assistance compression than the same assistance supplied afterward.
\end{proposition}

The proof is given in the Appendix. The proposition isolates temporal order from total exposure. Both schedules contain one assisted and one unassisted block, and both end with the same accumulated breadth. The difference arises because assistance interacts with the breadth available \emph{when stabilization occurs}. Under immediate assistance, $B$ remains at $B_0$ throughout the assisted block, so compression is generated at high susceptibility $1-B_0$. Under delayed assistance, independent exploration first raises $B$, reducing the compressive force of the later assisted block.

The condition $\rho>\delta$ is substantively important. Early stabilization does not mechanically dominate in every environment. If compressed structure relaxes very quickly relative to repertoire formation, the early compression has time to dissipate and the recency of delayed assistance can dominate. The theory therefore predicts where a timing effect should be found: tasks in which independent exploration rapidly generates alternatives while stabilized representations decay slowly.

\begin{figure}[!t]
\centering
\includegraphics[width=0.96\textwidth]{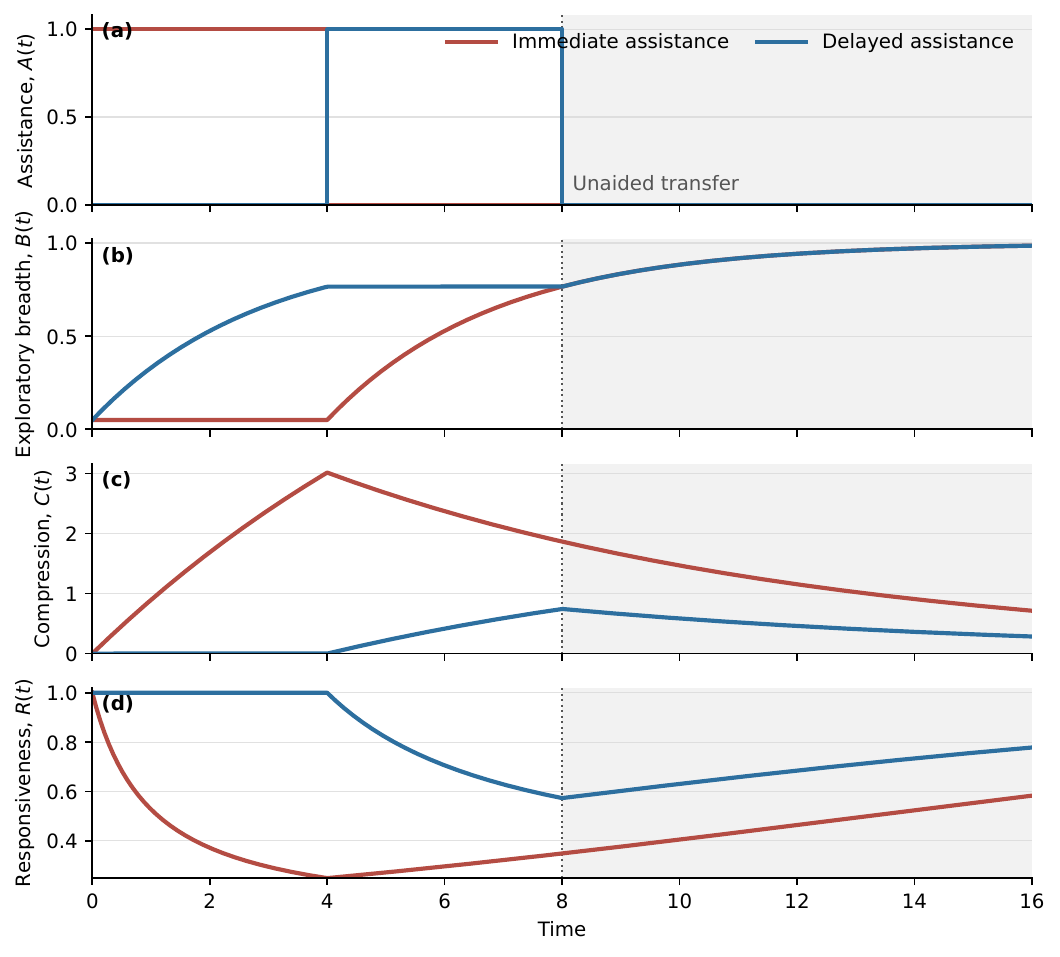}
\caption{Equal-dose immediate and delayed predictive assistance. The schedules contain the same assisted and unassisted durations. Delayed assistance allows exploratory breadth $B(t)$ to form before stabilization. When $\rho>\delta$, immediate assistance generates more compression $C(t)$ and lower exploratory responsiveness $R(t)$ at the common withdrawal time $2L$. After withdrawal, the gap narrows gradually as compression relaxes. The trajectories are illustrative and use the parameter values reported in the Appendix.}
\label{fig:timing}
\end{figure}

Because $R'(C)<0$, Equation~\eqref{eq:timingcondition} immediately implies
\begin{equation}
R_E(2L)<R_D(2L)\quad\text{when }\rho>\delta.
\end{equation}
Immediate assistance then leaves lower exploratory variance and entropy at the common post-training test despite identical assistance dose and identical accumulated breadth. Figure~\ref{fig:timing} illustrates this comparison for a parameterization satisfying $\rho>\delta$.

\subsection{Withdrawal and recovery}

Suppose all assistance is removed at $T=2L$. For either schedule $j\in\{E,D\}$,
\begin{equation}
C_j(T+s)=C_j(T)e^{-\delta s},\qquad s\geq0,
\label{eq:recovery}
\end{equation}
and responsiveness recovers according to
\begin{equation}
R_j(T+s)=\frac{\lambda_0}{\lambda_0+C_j(T)e^{-\delta s}}.
\end{equation}
Removal of assistance therefore does not instantaneously restore exploratory mobility. The relevant recovery scale is $1/\delta$, and the half-life of compression is $\log(2)/\delta$.

This persistence is a fading structural memory, not equilibrium hysteresis. For fixed assistance and breadth, Equation~\eqref{eq:compression} has a unique equilibrium, and after withdrawal both schedules eventually converge toward $C=0$. A loop obtained by plotting $C$ against a changing $A$ is therefore rate-dependent lag. Strong hysteresis, bistability, or irreversible collapse would require additional nonlinear feedbacks not assumed here.

The local geometry can nevertheless make finite-horizon recovery behaviorally consequential. For a fixed basin boundary $|X-m|=a$, the small-noise mean escape time associated with Equation~\eqref{eq:attention} has the leading exponential form
\begin{equation}
\mathbb{E}[\tau_{\mathrm{escape}}]\propto
\exp\!\left(\frac{[\lambda_0+C]a^2}{2D}\right),
\label{eq:escape}
\end{equation}
up to curvature-dependent prefactors. A moderate compression difference can therefore produce a much larger difference in the time required to leave a stabilized region during a finite task. Equation~\eqref{eq:escape} does not imply a universal collapse threshold; any operational threshold depends on the basin boundary, perturbation level, and observation horizon.

\section{Empirical Test}

The theory is organized around a single estimand: the effect of assistance timing on \emph{unaided exploratory behavior after matched assistance}. A suitable flagship experiment is a causal-discovery microworld in which the hypothesis and intervention spaces can be enumerated in advance. Participants learn the structure of a small system by proposing causal models and choosing informative experiments. The task should admit several plausible initial models, include evidence capable of disconfirming an attractive early hypothesis, and culminate in structurally related transfer problems.

Participants are randomly assigned to three conditions. In the immediate-assistance condition, a predictive system supplies a candidate causal model and suggested next experiment at the beginning of the learning episode. In the delayed-assistance condition, participants first complete an independent exploration block and then receive the identical candidate model, experimental suggestion, and total duration of support. A no-assistance condition identifies the unaided baseline. After training, all assistance is removed and participants complete multiple transfer trials, including at least one trial containing evidence that contradicts the assisted model.

The primary outcome is entropy over preregistered strategy or hypothesis classes during unaided transfer,
\begin{equation}
H=-\sum_i p_i\log p_i,
\end{equation}
where $p_i$ is the share of exploratory actions allocated to class $i$ \cite{shannon1948mathematical}. This behavioral entropy is not identified with differential entropy in Equation~\eqref{eq:entropy}; it is an observable proxy for the diversity of realized search. Secondary outcomes include the number of distinct causal models generated, coverage of the experiment space, latency to abandon the assisted hypothesis after contradiction, depth of model revision, transfer accuracy, and the recovery slope across successive unaided trials.

The timing mechanism makes four predictions. First, immediate assistance may improve or preserve training performance while reducing self-generated branching during the assisted episode. Second, when $\rho>\delta$, the immediate group should display lower exploratory entropy and slower switching than the delayed group during unaided transfer. Third, the between-group gap should contract gradually across transfer trials rather than disappear at the moment support is removed. Fourth, domains with rapid independent repertoire formation and slow decay of stabilized representations should exhibit the largest timing effect.

The experiment can also falsify the mechanism. The theory is not supported if matched immediate and delayed assistance produce indistinguishable post-withdrawal exploration, if any difference vanishes immediately upon removal, or if the estimated timing effect does not vary with independent indicators of repertoire formation and structural relaxation. Likewise, superior assisted-task performance alone does not confirm the theory because the claim concerns later unaided exploration.

Interaction traces provide natural measures of the latent states. Breadth $B$ can be approximated through the cumulative number or coverage of distinct self-generated hypothesis classes before guidance. Compression $C$ is more indirect and should be inferred from concentration around the assisted trajectory, resistance to contradictory evidence, local transition rates, and post-withdrawal recovery. Longitudinal designs can estimate $\delta$ from the decay of these effects across unaided trials. Such measurement keeps the geometric language tied to observable behavior rather than treating curvature as a directly observed psychological quantity.

\section{Design Implication and Boundary Conditions}

The direct design implication is temporal gating. In tasks where later independent exploration matters, a predictive system should ordinarily elicit an initial self-generated representation, hypothesis set, decomposition, or plan before stabilizing a candidate trajectory. The purpose is not to maximize difficulty or withhold useful information indefinitely. It is to ensure that assistance enters after enough independent breadth has formed to reduce concentrated stabilization around the first externally supplied path.

This implication can be implemented through delayed completion, generate-before-view interfaces, requirements to produce multiple candidate strategies before receiving a recommendation, or assistance triggered by evidence of exploratory coverage rather than elapsed time alone. Intermittent support may also help when unassisted intervals expand breadth, but the model does not imply that every pulsed schedule dominates every continuous schedule. The effect depends on how much breadth forms during the unassisted intervals and how quickly compression relaxes.

The same dynamics can favor immediate compression in other settings. Emergency response, high-reliability operations, and standardized procedures often prioritize rapid convergence on validated actions rather than broad search. In those tasks, a narrow and stable basin can be an intended outcome. Predictive assistance may also expand rather than compress exploration when it presents competing hypotheses, seeks disconfirming evidence, or enables traversal that the user could not otherwise perform. Such systems fall outside the positive-$A$ mechanism to the extent that they increase self-generated or jointly generated breadth rather than stabilize a single candidate path.

The normative question is therefore task-relative. Temporal gating is most relevant when independent transfer, revision after contradiction, innovation, or adaptation to novelty is part of the objective. Immediate stabilization is more defensible when errors from exploratory wandering are severe and later unaided flexibility has little value.

\section{Discussion}

The paper advances a dynamical theory of exploration under predictive optimization. Its contribution is not the general observation that people can rely on tools, nor the claim that externally supplied answers always reduce learning. It identifies a temporal interaction: the compressive effect of predictive assistance depends on the exploratory breadth present when stabilization occurs. This interaction distinguishes compression earned after exploration from compression imposed before exploration.

The model deliberately separates this cognitive mechanism from broader questions about labor substitution, social knowledge production, or the welfare consequences of decentralized AI adoption. Those questions require agents, incentives, and externalities. Here the unit of analysis is a deliberative trajectory, and the outcome is later unaided exploratory mobility. The narrow scope makes the timing mechanism experimentally identifiable.

Several limitations delimit the claim. The local quadratic landscape represents one focal basin and does not describe the full topology of a high-dimensional cognitive space. Exploratory breadth is summarized by a scalar rather than a structured repertoire, and the model does not distinguish useful alternatives from noise. Assistance is represented by its stabilizing effect rather than by detailed interface features. The model also conditions on the immediate usefulness of the supplied trajectory instead of explaining short-run performance. These simplifications permit an exact timing comparison, but empirical applications must specify strategy classes, basin boundaries, and the mapping from interaction traces to breadth and compression.

The recovery process is likewise intentionally conservative. Linear relaxation produces delayed but eventual recovery. Persistent multiple equilibria, irreversible loss, or critical collapse are not results of the present model. They may arise if compression reduces future breadth formation, if repeated exposure changes the relaxation rate, or if a finite developmental window closes before lost exploration can be recovered. Such extensions should be evaluated separately rather than inferred from the baseline equations.

The model's strongest implication is comparative. Equal amounts of assistance need not be developmentally or behaviorally equivalent. When exploratory breadth forms faster than stabilized structure relaxes, moving assistance from the beginning to a later point in the same learning episode reduces residual compression without reducing the assistance dose. This prediction connects theory directly to interface scheduling and provides a tractable target for empirical evaluation.

\section{Conclusion}

Predictive systems can supply a stable trajectory before a learner has generated alternatives. This paper asks what that temporal reversal does to later exploratory capacity. The answer is conditional but precise: assistance before diversification creates greater residual compression than equal assistance after diversification when repertoire formation is faster than structural relaxation. The consequence is lower post-assistance exploratory variance, lower strategy diversity, and slower recovery after support is withdrawn. Predictive assistance should therefore be evaluated not only by what it helps users produce now, but also by when stabilization enters the exploratory process and what independent search remains possible afterward.

\bibliographystyle{unsrt}
\bibliography{sugar}

\section*{Author Contributions}
\textbf{BB} designed the research, developed the model, and wrote the manuscript.

\section*{Acknowledgements and Disclosures}
The author declares no relevant financial relationships or other potential conflicts of interest.

\appendix

\section{Mathematical Derivations}

\subsection{Exploratory breadth}

For any measurable assistance path $A(t)\in[0,1]$, Equation~\eqref{eq:breadth} has the solution
\begin{equation}
B(t)=1-(1-B_0)\exp\!\left[-\rho\int_0^t(1-A(s))\,ds\right].
\label{eq:breadthsolution}
\end{equation}
Thus, over a fixed horizon, paths containing the same total unassisted time generate the same terminal breadth in the baseline model. This property allows the timing comparison to hold breadth and assistance dose fixed.

For both block schedules,
\begin{equation}
\int_0^{2L}[1-A_j(s)]\,ds=L,
\end{equation}
so Equation~\eqref{eq:breadthsolution} gives
\begin{equation}
B_j(2L)=1-(1-B_0)e^{-\rho L},\qquad j\in\{E,D\}.
\end{equation}

\subsection{Compression under the two schedules}

The general solution to Equation~\eqref{eq:compression} is
\begin{equation}
C(T)=C_0e^{-\delta T}+\alpha\int_0^T A(s)[1-B(s)]e^{-\delta(T-s)}\,ds.
\label{eq:compressionsolution}
\end{equation}

Under immediate assistance, $A_E=1$ and $B_E(t)=B_0$ on $[0,L]$. Compression at the end of the assisted block is
\begin{equation}
C_E(L)=\frac{\alpha(1-B_0)}{\delta}(1-e^{-\delta L}).
\end{equation}
Assistance is then removed, so compression decays for another interval of length $L$:
\begin{equation}
C_E(2L)=\frac{\alpha(1-B_0)}{\delta}(1-e^{-\delta L})e^{-\delta L}.
\end{equation}

Under delayed assistance, breadth first grows without compression. At $t=L$,
\begin{equation}
1-B_D(L)=(1-B_0)e^{-\rho L}.
\end{equation}
During the second block, $A_D=1$, so breadth remains fixed and compression satisfies a linear forced equation. Hence
\begin{equation}
C_D(2L)=\frac{\alpha(1-B_0)}{\delta}(1-e^{-\delta L})e^{-\rho L}.
\end{equation}
Taking the ratio gives
\begin{equation}
\frac{C_E(2L)}{C_D(2L)}=e^{(\rho-\delta)L},
\end{equation}
which proves Proposition~1.

\subsection{Recovery time}

Let $\bar R\in(0,1)$ be an operational responsiveness threshold. From Equation~\eqref{eq:responsiveness}, the corresponding compression threshold is
\begin{equation}
\bar C=\lambda_0\left(\frac{1}{\bar R}-1\right).
\end{equation}
If $C_j(2L)>\bar C$, the additional time required to reach $R\geq\bar R$ after common withdrawal is
\begin{equation}
T_j(\bar R)=\frac{1}{\delta}\log\!\left(\frac{C_j(2L)}{\bar C}\right).
\end{equation}
Whenever both schedules begin above the threshold,
\begin{equation}
T_E(\bar R)-T_D(\bar R)
=\frac{1}{\delta}\log\!\left(\frac{C_E(2L)}{C_D(2L)}\right)
=\frac{(\rho-\delta)L}{\delta}.
\end{equation}
The early schedule therefore requires more recovery time exactly when $\rho>\delta$.

\subsection{Local variance, entropy, and escape}

For frozen $C$, Equation~\eqref{eq:attention} has stationary distribution
\begin{equation}
X\mid C\sim\mathcal{N}\!\left(m,\frac{D}{\lambda_0+C}\right).
\end{equation}
Equations~\eqref{eq:variance}--\eqref{eq:entropy} follow immediately. For a fixed boundary $a$ from the basin center, the potential difference is
\begin{equation}
\Delta U(C)=U(m+a)-U(m)=\frac{1}{2}(\lambda_0+C)a^2.
\end{equation}
The leading small-noise first-passage scaling is proportional to $\exp[\Delta U(C)/D]$, yielding Equation~\eqref{eq:escape}. This approximation treats $C$ as slow relative to the local attentional dynamics.

\section{Simulation Details}

Figure~\ref{fig:timing} numerically integrates Equations~\eqref{eq:breadth} and \eqref{eq:compression} with
\begin{equation}
B_0=0.05,\quad C_0=0,\quad \rho=0.35,\quad \delta=0.12,
\quad \alpha=1,\quad \lambda_0=1,\quad D=0.25,\quad L=4.
\end{equation}
The immediate schedule sets $A=1$ on $[0,4)$ and $A=0$ thereafter. The delayed schedule sets $A=0$ on $[0,4)$, $A=1$ on $[4,8)$, and $A=0$ thereafter. The plot continues to $t=16$ to show recovery after the common withdrawal time $2L=8$. Parameter values are illustrative rather than calibrated. The qualitative timing result follows analytically from Proposition~1 and does not depend on this particular parameterization.

\end{document}